\documentclass[10pt,twocolumn,letterpaper]{article}

\usepackage{cvpr}              % To produce the CAMERA-READY version
\usepackage{graphicx}
\usepackage{amsmath}
\usepackage{amssymb}
\usepackage{booktabs}
\usepackage{times}
\usepackage{epsfig}
\usepackage{graphicx}
\usepackage{enumitem}  % For compact lists

\usepackage[export]{adjustbox}
\usepackage{array}
\usepackage{multirow}
\usepackage{tabularx}
\usepackage{support-caption}
\usepackage{subcaption}
\usepackage[utf8]{inputenc}
\usepackage[export]{adjustbox}
\usepackage{wrapfig}
\usepackage{threeparttable}
\usepackage[english]{babel}
\usepackage{color, colortbl}
\usepackage{makecell}

\usepackage{bm} 
\usepackage{dsfont}
\usepackage{arydshln}
\usepackage{comment}
\usepackage[pagebackref,breaklinks,colorlinks]{hyperref}

\usepackage[capitalize]{cleveref}
\crefname{section}{Sec.}{Secs.}
\Crefname{section}{Section}{Sections}
\Crefname{table}{Table}{Tables}
\crefname{table}{Tab.}{Tabs.}

\def\cvprPaperID{648} % *** Enter the CVPR Paper ID here
\def\confName{CVPR}
\def\confYear{2023}

\begin{document}

\title{TransCAR: Transformer-based Camera-And-Radar Fusion for 3D Object Detection}

\author{Su Pang, Daniel Morris, Hayder Radha\\
Michigan State University\\
%426 Auditorium Road, East Lansing, MI 48824\\
{\tt\small pangsu@msu.edu, dmorris@msu.edu, radha@msu.edu}
}
\maketitle

%%%%%%%%% ABSTRACT
\begin{abstract}
   Despite radar's popularity in the automotive industry, for fusion-based 3D object detection, most existing works focus on LiDAR and camera fusion. In this paper, we propose TransCAR, a Transformer-based Camera-And-Radar fusion solution for 3D object detection. Our TransCAR consists of two modules. The first module learns 2D features from surround-view camera images and then uses a sparse set of 3D object queries to index into these 2D features. The vision-updated queries then interact with each other via transformer self-attention layer. The second module learns radar features from multiple radar scans and then applies transformer decoder to learn the interactions between radar features and vision-updated queries. The cross-attention layer within the transformer decoder can adaptively learn the soft-association between the radar features and vision-updated queries instead of hard-association based on sensor calibration only. Finally, our model estimates a bounding box per query using set-to-set Hungarian loss, which enables the method to avoid non-maximum suppression. TransCAR improves the velocity estimation using the radar scans without temporal information. The superior experimental results of our TransCAR on the challenging nuScenes datasets illustrate that our TransCAR outperforms state-of-the-art Camera-Radar fusion-based 3D object detection approaches.
\end{abstract}

%%%%%%%%% BODY TEXT
\section{Introduction}
\label{sec:intro}

Radars have been used for Advanced Driving Assistance System (ADAS) for many years. However, despite radar's popularity in the automotive industry, when considering 3D object detection most existing works focus on LiDAR \cite{zhou2018voxelnet,yan2018second,shi2019pointrcnn,lang2019pointpillars,qi2018frustum,std2019yang,shi2020pv,yin2021center}, camera \cite{brazil2019m3d,wang2022detr3d,CaDDN,chen2020monopair} and LiDAR-camera fusion \cite{chen2017multi,ku2018joint,qi2018frustum,liang2018deep,wang2019frustum,huang2020epnet,yoo20203d,xie2020pi,liang2019multi,pang2020clocs,pang2022fast}. One reason for this is that there are not as many open datasets annotated with 3D bounding boxes that include radar data \cite{caesar2020nuscenes,chang2019argoverse,geiger2012we,sun2020scalability}. Another reason is that, compared to LiDAR point clouds, automotive radar signals are much sparser and lack height information. These properties make it challenging to distinguish between returns from objects of interest and backgrounds. However, radar has its strengths compared to LiDAR: (1) radar is robust under adverse weather and light conditions; (2) radar can accurately measure object's radial velocity through the Doppler effect without requiring temporal information from multiple frames; (3) radar has much lower cost compared to LiDAR.  Therefore, we believe there is a strong potential for performance gain by pursuing radar-camera fusion research.

\begin{figure}[t]
   \centering
    \includegraphics[width=\columnwidth]{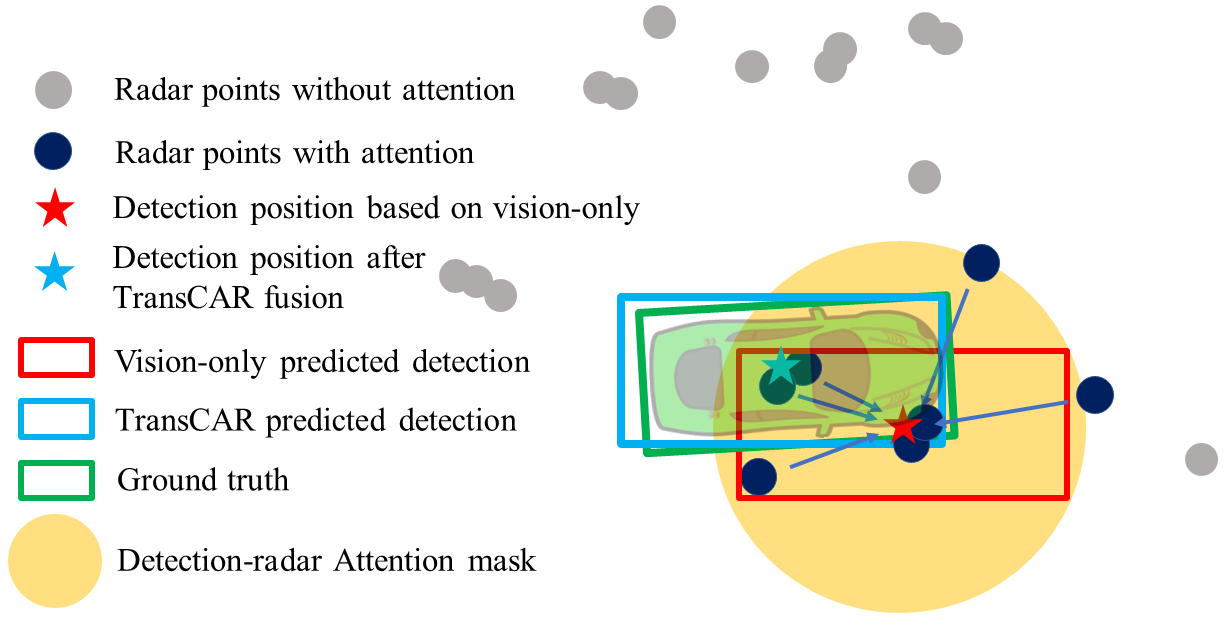}
    \centering
  \caption{An example from nuScenes \cite{caesar2020nuscenes} showing how TransCAR fusion works. Vision-only detection has significant range error. Our TransCAR fusion can learn the interactions between vision-based query and related radar signals and predict improved detection. Unrelated radar points are prevented from attention by Query-Radar attention mask.}
\label{main_idea}
\end{figure}

3D object detection is essential for self-driving and ADAS systems. The goal of 3D object detection is to predict a set of 3D bounding boxes and category labels for objects of interest. It is challenging to directly estimate and classify 3D bounding boxes from automotive radar data alone due to its sparsity and lack of height information. Monocular camera-based 3D detectors \cite{mousavian20173d,brazil2019m3d,chen2020monopair,CaDDN,wang2022detr3d} can classify objects, predict heading angles and azimuth angles of objects accurately. However, the errors in depth estimation are significant because regressing depth from a single image is inherently an ill-posed inverse problem. 
Radar can provide accurate depth measurement, which monocular camera-based solutions cannot. Camera can produce classification and 3D bounding box estimation that radar-based solutions cannot. Therefore, it is a natural idea to fuse radar and camera for better 3D object detection performance.

\begin{figure*}[t]
\centering
   \includegraphics[width=2.0\columnwidth]{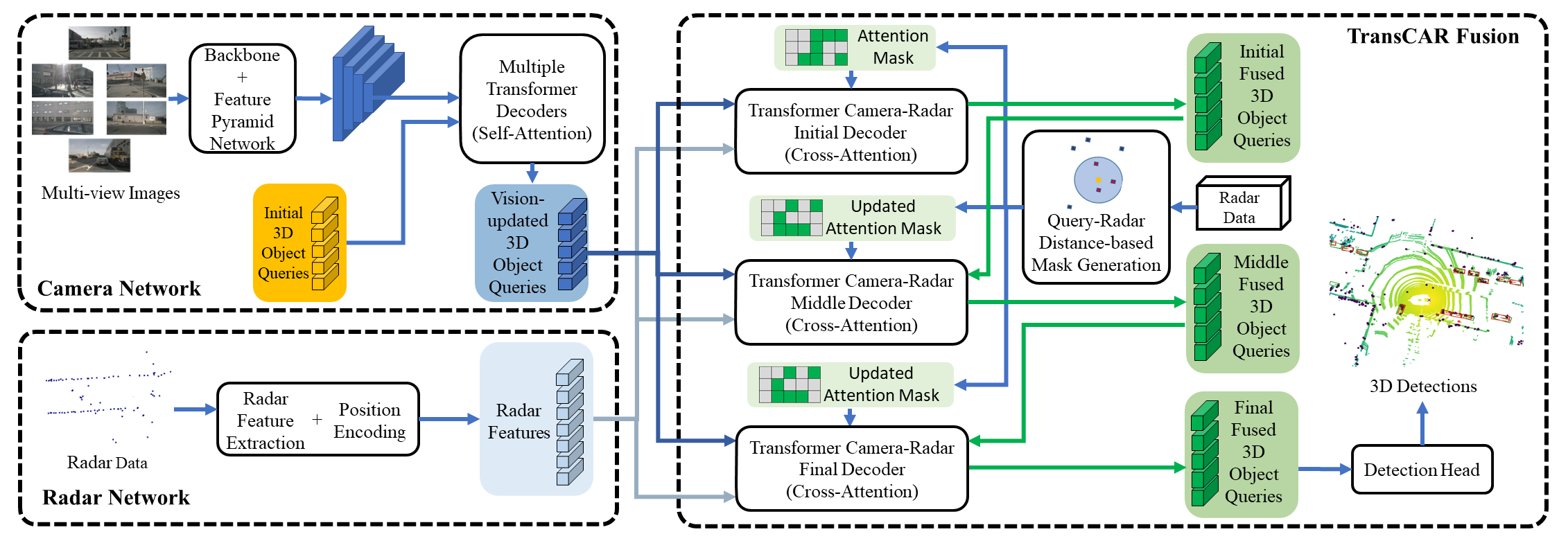}

   \caption{TransCAR system architecture. There are three primary components in the system: (1) A camera network (DETR3D\cite{wang2022detr3d}) based on transformer decoders to generate image-based 3D object queries. The initial object queries are generated randomly; (2) A radar network that encodes radar point locations and extracts radar features; (3) The TransCAR fusion module based on three transformer cross-attention decoders. We propose to use transformer to learn the interactions between radar features and vision-updated object queries for adaptive camera-radar association.} %Query-radar attention masks that prevent certain invalid attentions are applied to assist the network better attend the queries with closely related radar features.}
    
\label{system_architecture}
\end{figure*}

Data association between different sensor modalities is the main challenge for sensor fusion technologies. Existing works mainly rely on multi-sensor calibration to do pixel-level \cite{Vora_2020_CVPR}, feature level \cite{liang2019multi,liang2018deep,chen2017multi,ku2018joint,xu2018pointfusion} or detection level \cite{pang2020clocs,pang2022fast} association. However, this is challenging for radar and camera association. First, the lack of height measurement in radar makes the radar-camera projection incorporate large uncertainties along the height direction. Second, radar beams are much wider than a typical image pixel and can bounce around. This can result in some hits visible to the radar but are occluded from the camera. Third, radar measurements are sparse and have low resolution. Many objects visible to the camera do not have radar hits. For these reasons, the hard-coded data association based on sensor calibration performs poorly for radar and camera fusion.

The seminal \textit{Transformer} framework was initially proposed as a revolutionary technology for natural language processing (NLP) \cite{vaswani2017attention}, and subsequently has shown its versatility in computer vision applications including object classification \cite{dosovitskiy2020image} and detection \cite{carion2020end,zhu2020deformable}. The self-attention and cross-attention mechanism within the transformer can learn the interactions between multiple sets of information \cite{vaswani2017attention,carion2020end,bai2022transfusion}. And we believe that this makes the transformer framework a viable fit to solve the data association in camera-radar fusion. In this paper, we propose a novel Transformer-based Radar and Camera fusion network termed TransCAR to address the problems mentioned above. 

Our TransCAR first uses DETR3D \cite{wang2022detr3d} to generate image-based object queries. Then TransCAR learns radar features from multiple accumulated radar scans and applies a transformer decoder to learn the interactions between radar features and vision-updated queries. The cross-attention within the transformer decoder can adaptively learn the soft-association between the radar features and vision-updated queries instead of hard-association based on sensor calibration only. Finally, our model predicts a bounding box per query using a set-to-set Hungarian loss. Figure~\ref{main_idea} illustrates the main idea of TransCAR. We also add the velocity discrepancy as a metric for the Hungarian bipartite matching because radar can provide accurate radial velocity measurements. Although our focus is on fusing multiple monocular cameras and radars, the proposed TransCAR framework is applicable to stereo camera systems as well. We demonstrate our TransCAR using the challenging nuScenes dataset \cite{caesar2020nuscenes}. TransCAR outperforms all other state-of-the-art (SOTA) camera-radar fusion-based methods by a large margin. The proposed architecture delivers the following contributions:

\begin{itemize}

\item We study the characteristics of radar data and propose a novel camera-radar fusion network that adaptively learns the soft-association, and we show superior 3D detection performance compared to hard-association based on radar-camera calibration.
\item We propose a novel camera-radar fusion network that adaptively learns the soft-association, and we show superior 3D detection performance compared to hard-association based on radar-camera calibration.
\item The Query-Radar attention mask is proposed to assist the cross-attention layer to avoid unnecessary interactions between faraway vision queries and radar features, and in better learning the associations.
\item TransCAR improves the velocity estimation using radar without requiring temporal information.
\item At the time of submission, TransCAR ranks \textbf{1st} among published camera-radar fusion-based methods on the nuScenes 3D detection benchmark. %And we even outperforms some early released LiDAR solutions.

\end{itemize}
%-------------------------------------------------------------------------
\section{TransCAR}
A high-level diagram of the proposed TransCAR architecture is shown in Figure~\ref{system_architecture}. The camera network first utilizes surround-view images to generate vision-updated object queries. The radar network encodes radar point locations and extract radar features. Then the TransCAR fusion module fuses the vision-updated object queries with useful radar features. In the following, we present the details of each module in TransCAR.
%Then the TransCAR fusion module performs camera-radar fusion by attentively fusing vision-updated object queries with useful radar features. In the following, we present the details of each module in TransCAR.
%-------------------------------------------------------------------------
\subsection{Camera Network}
\label{camera network}

Our camera network takes surround-view images collected by 6 cameras covering the full 360 degrees around the ego-vehicle and initial 3D object queries as input, and outputs a set of vision-updated 3D object queries in the 3D space. We apply DETR3D \cite{wang2022detr3d} to the camera network and follow the iterative top-down design. It utilizes initial 3D queries to index 2D features for refining 3D queries. The output 3D vision-updated queries are the input for the TransCAR fusion module.

\subsubsection{Why Start from Camera}
We use surround-view images to generate 3D object queries for fusion. Radar is not suitable for this task because many objects of interest do not have radar returns. There are two main reasons behind this. First, a typical automotive radar has a very limited vertical field of view compared to both camera and LiDAR, and a it is usually installed at a lower position. Therefore, any object that is located out of the radar's small vertical field of view will be missed. Second, unlike LiDAR, the radar beams are wider and the azimuth resolution is limited, making it difficult to detect small objects. According to our statistics in supplementary materials, in the nuScenes training set, radar has a high miss rate, especially for small objects. For the two most common classes on the road, car and pedestrian, radar misses 36.05\% of cars and 78.16\% of pedestrians. Cameras have much better object visibilities. Therefore, we utilize images first to predict 3D object queries for fusion.

\subsubsection{Methodology}
The camera network uses ResNet-101 \cite{he2016deep} with Feature Pyramid Network (FPN) \cite{lin2017feature} to learn a multi-scale feature pyramid. These multi-scale feature maps provide rich information for detecting objects in different sizes. Following \cite{zhu2020deformable,wang2021object}, our camera network (DETR3D \cite{wang2022detr3d}) is iterative. It has 6 transformer decoder layers to produce vision-updated 3D object queries; each layer takes the output queries from the previous layer as input. The steps within each layer are explained below.

For the first decoder layer, a set of $N$ ($N=900$ for nuScenes) learnable 3D object queries $\bm{Q}^0=\{\mathbf{q}_1^0, \mathbf{q}_2^0,..., \mathbf{q}_N^0\}\in\mathbb{R}^C$ are initialized randomly within the 3D surveillance area. The superscript $0$ represents the input query to the first layer, and the subscript is the index of the query. The network learns the distribution of these 3D query positions from the training data. For the following layers, the input queries are the output queries from the previous layer. Each 3D object query encodes a 3D center location $\mathbf{p}_i\in\mathbb{R}^3$ of a potential object. These 3D center points are projected to the image feature pyramid based on the camera extrinsic and intrinsic parameters to sample image features via bilinear interpolation. Assuming there are $k$ layers in the image feature pyramid, the sampled image feature $\bm{f}_i\in\mathbb{R}^C$ for a 3D point $\mathbf{p}_i$ is the sum of sampled features across all $k$ levels, $C$ is the number of feature channels. A given 3D center point $\mathbf{p}_i$ may not be visible in any camera image. We pad the sampled image features corresponding to these out-of-view points with zeros. 

A Transformer self-attention layer is used to learn the interactions among $N$ 3D object queries and generate attention scores. The object queries are then combined with the sampled image features weighted by the attention scores to form the updated object queries $\bm{Q}^l=\{\mathbf{q}_1^l, \mathbf{q}_2^l,..., \mathbf{q}_N^l\}\in\mathbb{R}^C$, where $l$ is the current layer. $\bm{Q}^l$ is the input set of queries for the $(l+1)$-th layer.

For each updated object query $\mathbf{q}_i^l$, a 3D bounding box and a class label are predicted using two neural networks. The details of bounding box encoding and loss function are described in Section \ref{box and loss}. A loss is computed after each layer during training. In inference mode, only the vision-updated queries output from the last layer are used for fusion.

\subsection{Radar Network}
\label{section_radar_network}
The radar network is designed to learn useful radar features and encode their 3D positions for fusion. We first filter radar points according to $x$ and $y$ range, since only objects within $+/-50$ meters box area in BEV are evaluated in nuScenes \cite{caesar2020nuscenes}. As radar is sparse, we accumulate radar from the previous 5 frames and transform them into the current frame. The nuScenes dataset provides 18 channels for each radar point, including the 3D location $x,y,z$ in ego vehicle frame, radial velocities $v_x$ and $v_y$, ego vehicle motion compensated velocities $v_{xc}$ and $v_{yc}$, false alarm probability $pdh0$, a dynamic property channel $dynProp$ indicating whether or not the cluster is moving or stationary, and other state channels \footnote{A detailed explanation of each channel can be found at: {https://github.com/nutonomy/nuscenes-devkit/blob/master/python-sdk/nuscenes/utils/data\_classes.py}.}. To make the state channels feasible for the network to learn, we transform them into one-hot vectors. Since we use 5 accumulated frames, the time offset of each frame with regard to the current timestamp is useful to indicate the position offset, so we also add a time offset channel for each point. With these pre-processing operations, each input radar point has 36 channels.

\begin{figure}[t]
\centering
   \includegraphics[width=0.95\columnwidth]{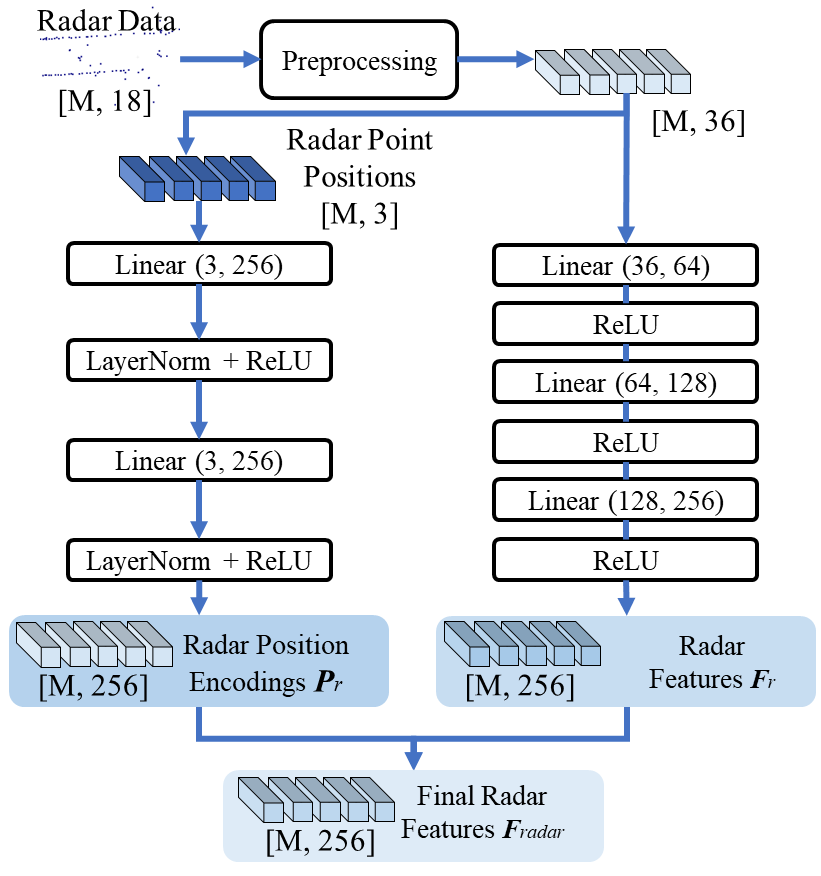}

   \caption{Details of radar network. The position encoding network (left) takes radar point positions ($xyz$) as input. The radar data after preprocessing (Section \ref{section_radar_network}) are sent to the radar feature extraction network (right) to learn useful radar features. Since radar signal is very sparse, each radar point is treated independently. The numbers within the square brackets represent the shape of the data. }
    
\label{radar_network}
\end{figure}

Multilayer perceptron (MLP) networks are used to learn radar features $\bm{F}_r\in\mathbb{R}^{M\times C}$ and radar point position encodings $\bm{P}_r\in\mathbb{R}^{M\times C}$, where $M$ and $C$ are the number of radar points and the number of feature channels, respectively. In this paper, we set $M=1500$ and $C=256$ for nuScenes dataset. Note that there are less than 1500 radar points for each timestep even after accumulation in nuScenes dataset. Therefore, we pad the empty spots with out-of-scope positions and zero features for dimension compatibility. Figure \ref{radar_network} shows the details of the radar network. We combine the learned features and position encodings as the final radar features $\bm{F}_{radar}=(\bm{F}_r + \bm{P}_r)\in\mathbb{R}^{M\times C}$. These final radar features together with the vision-updated queries from the camera network are used for TransCAR fusion in the next step.

\subsection{TransCAR Fusion}
TransCAR fusion module takes vision-updated queries and radar features from previous steps as input, and outputs fused queries for 3D bounding box prediction. Three transformer decoders work in an iterative fashion in the TransCAR fusion module. The query-radar attention mask is proposed to assist the cross-attention layer in better learning the interactions and associations between vision-updated queries and radar features.

\begin{figure}[t]
\centering
   \includegraphics[width=0.45\textwidth]{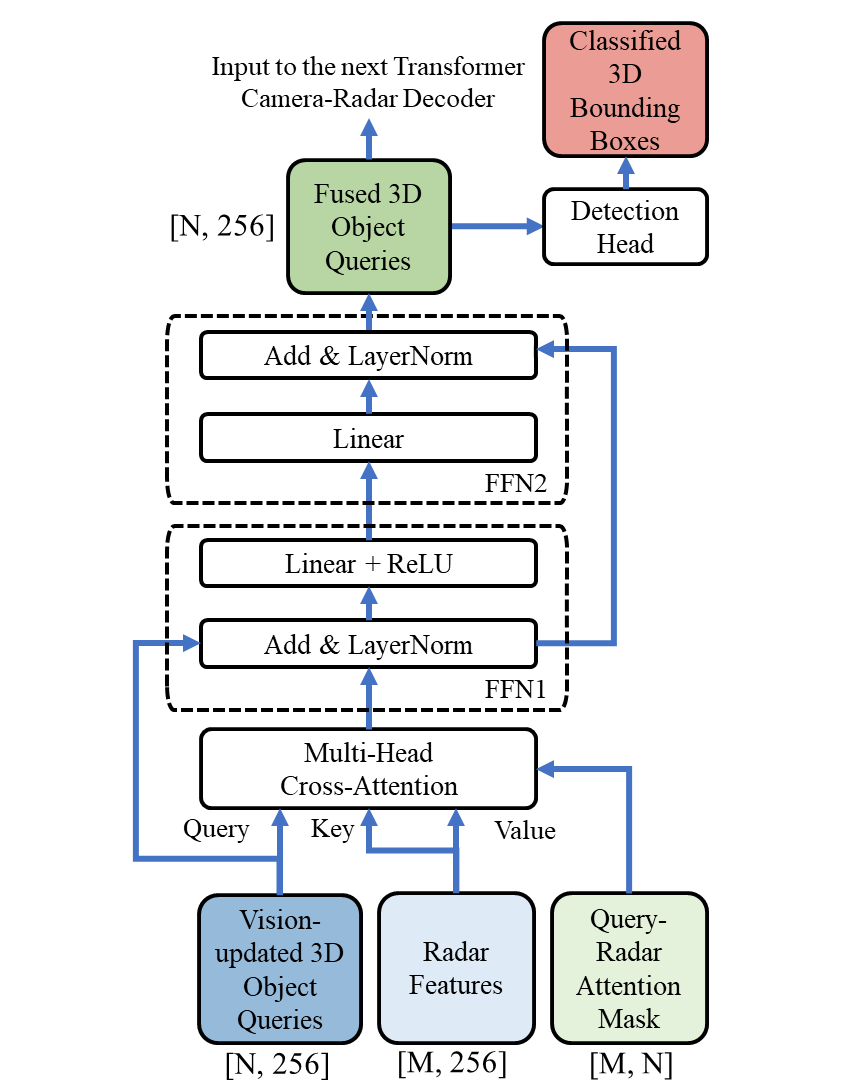}

   \caption{Details of transformer camera-radar decoder layer. The vision-updated 3D object queries are the queries to the multi-head cross attention module. The radar features are keys and values. See Section \ref{TCRCA} for details. The numbers within the square brackets represent the shape of the data.}
\label{TransCAR_fusion_network}
\end{figure}

\subsubsection{Query-Radar Attention Mask}
It is challenging and time consuming to train a transformer if both the number of input queries, keys and values are large \cite{dosovitskiy2020image,carion2020end}. For our transformer decoder, there are $N$ 3D object queries $\bm{Q}\in\mathbb{R}^{N\times C}$ and $M$ radar features $\bm{F}_{rad}\in\mathbb{R}^{M\times C}$ as keys and values, where $N=900$ and $M=1500$ for nuScenes. It is not necessary to learn every pairwise interaction $(900\times 1500)$ between them. For a query $\mathbf{q_i}\in\bm{Q}$, only the nearby radar features are useful. There is no need to interact $\mathbf{q}_i$ with other radar features that are far away. Therefore, we define a binary $N\times M$ Query-Radar attention mask $\bm{M}\in\{0, 1\}^{N\times M}$ to prevent attention for certain positions, where $0$ indicates no attention and $1$ represents allowed attention. A position $(i,j)$ in $\bm{M}$ is allowed for attention only when the $xy$ Euclidean distance between the $i$-th query $\mathbf{q}_i$ and the $j$-th radar feature $\mathbf{f}_j$ is less than a threshold. There are three Query-Radar attention masks in TransCAR fusion corresponding to the three transformer decoders. The radii for these three masks are $2m$, $2m$ and $1m$, respectively.

\subsubsection{Transformer Camera and Radar Cross-Attention}
\label{TCRCA}
Three transformer cross-attention decoders are cascaded to learn the associations between vision-updated queries and radar features in our TransCAR fusion. Figure \ref{TransCAR_fusion_network} shows the details of one transformer cross-attention decoder. For the initial decoder, the vision-updated queries $\bm{Q}_{img}\in\mathbb{R}^{N\times C}$ output from the camera network are the input queries. The radar features $\bm{F}_{rad}\in\mathbb{R}^{M\times C}$ are the input keys and values. The Query-Radar attention mask $\bm{M}_1$ is used to prevent attentions to certain unnecessary pairs. The cross-attention layer within the decoder will output an attention score matrix $\mathbf{A}_1\in[0,1]^{N\times M}$. For the $M$ elements in the $i$-th row of $\mathbf{A}_1$, they represent the attention scores between the $i$-th vision-updated query and all $M$ radar features, and their sum is 1. Note that for each query, only radar features close to it are allowed for attention, so for each row in $\mathbf{A}_1$, most of them are zeros. These attention scores are indicators of associations between vision-updated queries and radar features. Then, the attention-weighted radar features for vision-updated queries are calculated as $\bm{F}_{rad1}^{\ast}=(\mathbf{A}_1 \cdot\bm{F}_{rad})  \in\mathbb{R}^{N\times C}$. These weighted radar features combined with the original vision-updated queries are then augmented by a feed-forward network (FFN) $\Phi_{FFN1}$. This forms the fused queries for initial-stage: $\bm{Q}_{f1}=\Phi_{FFN1}(\bm{Q}_{img} + \bm{F}_{rad1}^{\ast})\in\mathbb{R}^{N\times C}$.

The middle and final transformer decoders work similarly to the initial one. But they take the previous fused queries $\bm{Q}_{f1}$ instead of the vision-updated queries as input. Taking the middle query as an examole, the new Query-Radar attention mask $\bm{M}_2$ is calculated based on the distance between $\bm{Q}_{f1}$ and radar point positions. We also re-sample image features $\bm{f}_{f2}$ using encoded query positions in $\bm{Q}_{f1}$ as the query positions are updated in the initial decoder. Similarly to the initial decoder, the attention-weighted radar features for $\bm{Q}_{f1}$ are defined as $\bm{F}_{rad2}^{\ast}=(\mathbf{A}_2 \cdot\bm{F}_{rad})  \in\mathbb{R}^{N\times C}$, where $\mathbf{A}_2$ includes the attention scores for intial-stage fused query $\bm{Q}_{f1}$ and radar features $\bm{F}_{rad}$. The output fused queries are learned via  $\bm{Q}_{f2}=\Phi_{FFN2}(\bm{Q}_{f1} + \bm{F}_{rad2}^{\ast} + \bm{f}_{f2} )\in\mathbb{R}^{N\times C}$. We apply two sets of FFNs after the two decoders to perform bounding box predictions. We compute losses from the two decoders during training, and only the bounding boxes output from the last decoder are used during inference. 

Due visibility limiations, some queries may have no nearby radar signals. These queries will not interact with any radar signals, and their attention scores are all zeros. Detections from these queries will be vision-based only.

\subsection{Box Encoding and Loss Function}
\label{box and loss}
\textbf{Box Encoding:} We encode a 3D bounding box $\bm{b}_{3D}$ as an 11-digit vector:
\begin{align}
\bm{b}_{3D}=[\bm{cls},x,y,z,h,w,l,sin(\theta),cos(\theta),v_x,v_y]
\end{align}
where $\bm{cls}=\{c_1,...,c_n\}$ is the class label, $x,y$ and $z$ are the 3D center location, $h,w$ and $l$ are the 3D dimension, $\theta$ is the heading angle, $v_x$ and $v_y$ are the velocities along the $x$ and $y$ axes. For each output object query $\mathbf{q}$, the network predicts its class scores $\bm{c}\in[0,1]^{n}$ ($n$ is the number of classes, $n=10$ for nuScenes) and 3D bounding box parameters $\bm{b}\in\mathbb{R}^{10}$:
\begin{align}
\begin{split}
\bm{b}=[\Delta x,\Delta y,\Delta z,\log{h},\log{w},\log{l},\\sin(\theta),cos(\theta),v_x,v_y]
\end{split}
\end{align}
where $\Delta x,\Delta y$ and $\Delta z$ are the offsets between predictions and query positions from the previous layer. Different from DETR3D that estimates position offsets in the Sigmoid space \cite{wang2022detr3d}, we directly regress the positon offsets in the 3D Cartesian coordinates. DETR3D uses Sigmoid space because they want to keep the position outputs between $[0,1]$, so all the queries positions are within the distance boundaries. While for TransCAR, we started from optimized vision-updated queries whose positions are relatively more accurate. Therefore, we can avoid the redundant non-linear activations which could potentially impact the learning.

\textbf{Loss:} We use a set-to-set Hungarian loss to guide training and measure the difference between network predictions and ground truths following \cite{stewart2016end,carion2020end,wang2022detr3d}. There are two components in the loss function, one for classification and the other for bounding box regression. We apply focal loss \cite{lin2017focal} for classification to address the class imbalance, and $L1$ loss for bounding box regression. Assuming that $N$ and $K$ represent the number of predictions and ground truths in one frame, we pad $\phi$ (no object) with ground truths set since $N$ is significantly larger than $K$. Following \cite{stewart2016end,carion2020end,wang2022detr3d}, we use Hungarian algorithm \cite{kuhn1955hungarian} to solve the bipartite matching problem between the predictions and ground truths: 
\begin{multline}
\hat{\sigma} = \underset{\sigma\in\Theta}{arg\, min} \sum_{i}^{N} [ -\mathds{1}_{\{c_i \neq \phi\}} \hat{p}_{\sigma(i)}(c_i) + \\\mathds{1}_{\{c_i\neq \phi\}} \mathcal{L}_{box}(\mathbf{b}_i,\hat{\mathbf{b}}_{\sigma(i)})]
\end{multline}
where $\Theta$ denotes the set of permutations,  $\hat{p}_{\sigma(i)}(c_i)$ represents the probability of class $c_i$ with permutation index $\sigma(i)$, and $\mathcal{L}_{box}$ is the $L1$ difference for bounding boxes, $\mathbf{b}_i$ and $\hat{\mathbf{b}}_{\sigma(i)}$ are the ground truth box and predicted box respectively. Here, note that we also incorporate the velocity estimation $v_x$ and $v_y$ into $\mathcal{L}_{box}$ for a better match and velocity estimation. With the optimal permutation $\hat{\sigma}$, the final Hungarian loss can be represented as follows:
\begin{multline}
\mathcal{L}_{Hungarian} = \sum_{i}^{N} [ -\alpha(1-\hat{p}_{\hat{\sigma}(i)}(c_i))^{\gamma}\log\hat{p}_{\hat{\sigma}(i)}(c_i) + \\\mathds{1}_{\{c_i\neq \phi\}} \mathcal{L}_{box}(\mathbf{b}_i,\hat{\mathbf{b}}_{\hat{\sigma}(i)})]
\end{multline}
where $\alpha$ and $\gamma$ are the parameters of focal loss.
% \vspace{-0.25cm}
\section{Experimental Results}
We evaluate our TransCAR on the challenging nuScenes 3D detection benchmark \cite{caesar2020nuscenes} as it is the only open large-scale annotated dataset that includes radar.

\subsection{Dataset}
There are 6 cameras, 5 radars and 1 LiDAR installed on the nuScenes data collection vehicle. The nuScenes 3D detection dataset contains 1000 driving segments (scenes) of 20 seconds each, with 700, 150 and 150 segments for training, validation and testing, respectively. The annotation rate is 2Hz, so there are 28k, 6k and 6k annotated frames for training, validation and testing, respectively. There are 10 classes of objects. The true positive metric is based on BEV center distance.

\subsection{Evaluation Results}
We present our 3D detection results on the nuScenes test set in Table \ref{muscenes_test_table_version2}. Our TransCAR outperforms all other camera-radar fusion methods at the time of submission. Compared to the baseline camera-only method, DETR3D \cite{wang2022detr3d}, TransCAR has higher mAP and NDS (nuScenes Detection Score \cite{caesar2020nuscenes}). Noted that DETR3D is trained with CBGS \cite{zhu2019class}, while TransCAR is not. As shown in Table \ref{muscenes_test_table_version2}, among the 10 classes, the car class has the largest improvement ($+2.4\%$). Cars and pedestrians are the main objects of interest in driving scenarios. In the nuScenes dataset, class Car has the highest proportion in the training set, it accounts for $43.74\%$ of the total instances, and $63.95\%$ of these car instances have radar hits. Therefore, these car examples provide sufficient training examples for our TransCAR to learn the fusion. Class Pedestrian has the second highest proportion in the training set, it accounts for $19.67\%$ of the total instances, but only $21.84\%$ have radar returns. TransCAR can still improve pedestrian mAP by $1.2\%$. This demonstrates that, for objects with radar hits, TransCAR can leverage the radar hits to improve the detection performance, and for objects without radar hits, TransCAR can preserve the baseline performance.

Table \ref{nuscenes_car_ap} shows the quantitative comparison with baseline DETR3D \cite{wang2022detr3d} in Car class with different center distance evaluation metrics. In nuScenes dataset, the true positive metric is based on the center distance, which means the center distance between a true positive and the ground truth should be smaller than the threshold. nuScenes defines four distance thresholds ranging from 0.5 to 4.0 meters. As shown in Table \ref{nuscenes_car_ap}, TransCAR improves the AP for all 4 metrics. In particular, for the more strict and important metrics 0.5 and 1.0 meters thresholds, the improvement is $5.84\%$ and $6.19\%$ respectively.

\definecolor{LightCyan}{RGB}{153,204,255}
\definecolor{gg}{RGB}{0,153,76}
\definecolor{bb}{RGB}{0,0,204}
\begin{table*}[t]
\resizebox{\textwidth}{!}{
\setlength{\tabcolsep}{2.0pt} %1.5
\renewcommand{\arraystretch}{0.5}  %0.5
%\begin{center}

\begin{tabular}{|c|c|c|c|c|c|c|c|c|c|c|c|c|c|c|}
\hline
\hline
{\tiny{Method}} & {\tiny{Sensor}} & {\tiny NDS$\uparrow$} & {\tiny mAP$\uparrow$} & {\tiny mAVE$\downarrow$} & {\tiny Car} & {\tiny Truck} & {\tiny Bus} & {\tiny Trailer} & {\tiny C.V.} & {\tiny Ped.} & {\tiny Motor.} & {\tiny Bike} & {\tiny T.C.} & {\tiny Barrier}\\ 
\hline
{\tiny MonoDIS\cite{simonelli2019disentangling}}& {\tiny C}  & \tiny{38.4} & \tiny{30.4} & \tiny{1.553} & \tiny{47.8} & \tiny{22.0} & \tiny{18.8} & \tiny{17.6} & \tiny{7.4} & \tiny{37.0} & \tiny{29.0} & \tiny{24.5} & \tiny{48.7} & \tiny{51.1}\\
\hline
{\tiny CenterNet\cite{zhou2019objects}}& {\tiny C} & \tiny{40.0} & \tiny{33.8}  & \tiny{1.629} & \tiny{53.6} & \tiny{27.0} & \tiny{24.8} & \tiny{25.1} & \tiny{8.6} & \tiny{37.5} & \tiny{29.1} & \tiny{20.7} & \tiny{58.3} & \tiny{53.3} \\
\hline
{\tiny FCOS3D\cite{wang2021fcos3d} } & {\tiny C} & \tiny{42.8} & \tiny{35.8}  & \tiny{1.434} & \tiny{52.4} & \tiny{27.0} & \tiny{27.7} & \tiny{25.5} & \tiny{11.7} & \tiny{39.7} & \tiny{34.5} & \tiny{29.8} & \tiny{55.7} & \tiny{53.8} \\
\hline
{\tiny PGD\cite{wang2022probabilistic}} & {\tiny C} & \tiny{44.8} & \tiny{38.6}  & \tiny{1.509} & \tiny{56.1} & \tiny{29.9} & \tiny{28.5} & \tiny{26.6} & \tiny{13.4} & \tiny{44.1} & \tiny{39.7} & \tiny{31.4} & \tiny{60.5} & \tiny{56.1} \\
\hline
{\tiny DETR3D\cite{wang2022detr3d} (baseline)} & {\tiny C} & \tiny{47.9}  & \tiny{41.2} & \tiny{0.845} & \tiny{60.3} & \tiny{33.3} & \tiny{29.0} & \tiny{35.8} & \tiny{17.0} & \tiny{45.5} & \tiny{41.3} & \tiny{30.8} & \tiny{62.7} & \tiny{56.5} \\
\specialrule{1pt}{0pt}{0pt}
%\hdashline
{ \tiny  PointPillar \cite{lang2019pointpillars}}& {\tiny L}  & \tiny{45.3} & \tiny{30.5} & \tiny{0.316} & \tiny{68.4} & \tiny{ 23.0} & \tiny{28.2} & \tiny{23.4} & \tiny{4.1} & \tiny{59.7} & \tiny{27.4} & \tiny{1.1} & \tiny{30.8} & \tiny{38.9} \\
\hline
{\tiny infoFocus\cite{wang2020infofocus}}& {\tiny L}  & \tiny{39.5} & \tiny{39.5} & \tiny{1.000} & \tiny{77.9} & \tiny{31.4} & \tiny{44.8} & \tiny{37.3} & \tiny{10.7} & \tiny{63.4} & \tiny{29.0} & \tiny{6.1} & \tiny{46.5} & \tiny{47.8} \\
%\hline
%{  PointPainting \cite{Vora_2020_CVPR}}& {CL} & {  40.1} & { 55.0} & { 0.392} & { 0.270} & { 0.269} & { 0.476} & { 0.270} \\
%\hdashline
\specialrule{1pt}{0pt}{0pt}
{\tiny CenterFusion\cite{nabati2021centerfusion}} & {\tiny CR}  & \tiny{44.9} & \tiny{32.6} & \tiny{0.614} & \tiny{50.9} & \tiny{25.8} & \tiny{23.4} & \tiny{23.5} & \tiny{7.7} & \tiny{37.0} & \tiny{31.4} & \tiny{20.1} & \tiny{57.5} & \tiny{48.4} \\
\hline
{\tiny TransCAR(Ours)} & {\tiny CR}  & \tiny{\textbf{52.2}} & \tiny{\textbf{42.2}} &  \tiny{\textbf{0.495}} & \tiny{\textbf{62.7}} & \tiny{\textbf{33.6}} & \tiny{\textbf{30.0}} & \tiny{\textbf{36.0}} & \tiny{\textbf{17.9}} & \tiny{\textbf{46.7}} & \tiny{\textbf{43.1}} & \tiny{\textbf{32.2}} & \tiny{\textbf{62.9}} & \tiny{\textbf{57.0}} \\

\hline
\hline
\end{tabular}
%\end{center}
}
\caption{Quantitative comparison with SOTA methods on nuScenes test set. In `Sensor' colum, `C', `L' and `CR' represent camera, LiDAR, and camera-radar fusion, respectively. `C.V.', `Ped',`Motor' and `T.C' are short for construction vehicle, pedestrian, motorcycle, and traffic cone, respectively. TransCAR is currently the best camera-radar fusion-based method with the highest NDS and mAP, and it even outperforms early-released LiDAR-based approaches. The best performers are highlighted in bold, excluding LiDAR-only solutions.}
% \vspace{-0.5cm}
\label{muscenes_test_table_version2}
\end{table*}

\begin{table}[b]
%\resizebox{\textwidth}{!}{
\setlength{\tabcolsep}{1.5pt}
\renewcommand{\arraystretch}{1.1} 
\begin{center}

\begin{tabular}{|c|c|c|c|c|}
\hline
\hline
{\small Methods}  & \makecell{\small AP Car\\@0.5m} & \makecell{\small AP Car\\@1.0m} & \makecell{\small AP Car\\@2.0m} & \makecell{\small AP Car\\@4.0m}\\ 

\hline
\makecell{\small{Baseline(DETR3D)}}  & \small{16.72} & \small{46.28} & \small{71.55} & \small{83.96} \\%& {55.9} & {29.4} & {34.8} & \textbf{\color{gg}16.7} & {8.6} & {42.6} & {34.1} & {28.3} & {52.8} & \textbf{\color{gg}47.0} \\
\hline
{\small{TransCAR}}  & \small{22.56} & \small{52.47} & \small{74.52} & \small{84.52} \\
\hline
%\rowcolor{LightCyan}
{\small{Improvement}}  & \small{+5.84} & \small{+6.19} & \small{+2.97} & \small{+0.56} \\

\hline
\hline
\end{tabular}
\end{center}
\caption{Average Precision (AP) comparison with baseline DETR3D in Car class with different center-distance evaluation metrics on nuScenes validation set. Our TransCAR improves the AP by a large margin in all evaluation metrics.}
\label{nuscenes_car_ap}
\end{table}

\begin{table}
%\resizebox{\textwidth}{!}{
\setlength{\tabcolsep}{1.5pt}
\renewcommand{\arraystretch}{1.1} 
\begin{center}

\begin{tabular}{|c|c|c|c|c|}
\hline
\hline
{\small Method}  & {\small mAP$\uparrow$} & {\small NDS$\uparrow$} & {\small mATE$\downarrow$} & {\small mAVE$\downarrow$}\\ %& {\small Car} & {\small Truck} & {\small Bus} & {\small Trailer} & {\small C.V.} & {\small Ped.} & {\small Motor.} & {\small Bike} & {\small T.C.} & {\small Barrier}\\ 
\hline
{\small{Vision-only Baseline \cite{wang2022detr3d}}}  & \small{34.6} & \small{42.2} & \small{0.823} & \small{0.876} \\
\hline
\makecell{\small{w/o radar}\\\small{feature extraction}}  & \small{34.7} & \small{41.7} & \small{0.766} & \small{0.906} \\
\hline
\makecell{\small{w/o Query-Radar}\\\small{attention mask}}  & \small{34.4} & \small{39.5} & \small{0.765} & \small{1.125} \\
\hline
{\small{TransCAR(Ours)}}  & \small{\textbf{35.5}} & \small{\textbf{46.4}} & \small{\textbf {0.759}} & \small{\textbf{0.523}} \\
\hline
\hline
\end{tabular}
\end{center}
\caption{Ablation of the proposed TransCAR components on nuScenes val set.}
\label{nuscenes_ablation_table}
\end{table}

\begin{table}
%\resizebox{\textwidth}{!}{
\setlength{\tabcolsep}{1.0pt}
\renewcommand{\arraystretch}{1.1} 
\begin{center}

\begin{tabular}{|c|c|c|c|c|}

\hline
\small\makecell{Number of Transformer \\ Decoders in TransCAR}  & {\small mAP$\uparrow$} & {\small NDS$\uparrow$} & {\small mATE$\downarrow$} & {\small mAVE$\downarrow$}\\ 
\hline
\makecell{\small0 (Baseline,\\ \small without fusion } & {\small34.6} & {\small42.2} & {\small0.823} & {\small 0.876}\\
\hline
\makecell{\small{1}}  & \small{34.9} & \small{43.4} & \small{0.763} & \small{0.768} \\%& {55.9} & {29.4} & {34.8} & \textbf{\color{gg}16.7} & {8.6} & {42.6} & {34.1} & {28.3} & {52.8} & \textbf{\color{gg}47.0} \\
\hline
{\small{2}}  & \small{35.4} & \small{45.4} & \small{0.763} & \small{0.585} \\
\hline
{\small{3}}  & \small{\textbf{35.5}} & \small{\textbf{46.4}} & \small{\textbf{0.759}} & \small{\textbf{0.523}} \\

\hline
\hline
\end{tabular}
\end{center}
\caption{Evaluation on detection results from different number of transformer decoders in TransCAR.}
% \vspace{-0.25cm}
\label{nuscenes_fusion_layer_table}
\end{table}

\subsection{Qualitative Results}
\begin{figure*}
\centering
   \includegraphics[width=0.95\textwidth]{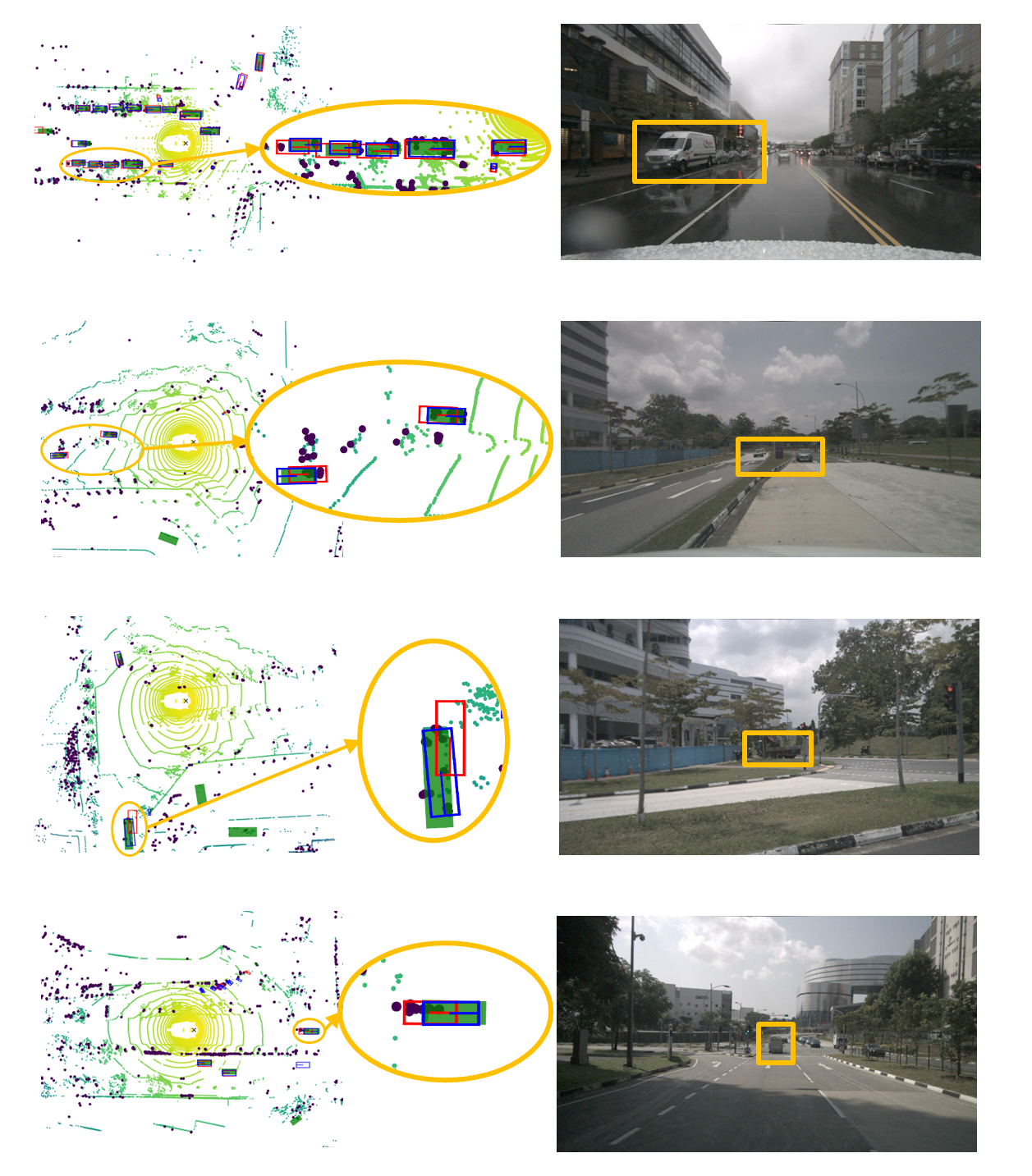}

   \caption{Qualitative comparison between TransCAR and baseline DETR3D on the nuScenes dataset \cite{caesar2020nuscenes}. Blue and red boxes are the predictions from TransCAR and DETR3D respectively, green filled rectangles are ground truths. The larger dark points are radar points, smaller color points are LiDAR points for reference (color yallow to green indicates the increasing distance).  The oval regions on the left column highlight the improvements made by TransCAR, the orange boxes on the image highlight the corresponding oval region in the top-down view. Best viewed with zoom-in and color.}
\label{qualitative}
\end{figure*}

Figure \ref{qualitative} shows the qualitative comparison between TransCAR and baseline DETR3D~\cite{wang2022detr3d} on the nuScenes dataset \cite{caesar2020nuscenes}. Blue and red boxes are the predictions from TransCAR and DETR3D respectively, green filled rectangles are ground truths. The larger dark points are radar points, smaller color points are LiDAR points for reference (color yallow to green indicates the increasing distance).  The oval regions on the left column highlight the improvements made by TransCAR, the orange boxes on the image highlight the corresponding oval region in the top-down view. TransCAR can fuse the detections from baseline DETR3D and improve the 3D bounding box estimation significantly.

\subsection{Ablation and Analysis}

Due to space constraints, we show part of the ablation studies in this section, more ablation studies are shown in the supplementary materials.

\textbf{Contribution of each component:} We evaluate the contribution of each component within our TransCAR network. The ablation study results on nuScenes validation set are shown in Table \ref{nuscenes_ablation_table}. The vision-only baseline is DETR3D \cite{wang2022detr3d}. Radial velocity is one of the unique measurements that radar can provide; although it is not true velocity, it can still guide the network to predict the object's velocity without temporal information. As shown in the second row of Table \ref{nuscenes_ablation_table}, without radar radial velocity, the network can only use the location of radar points for fusion, and the mAVE ($m/s$) is significantly higher (0.906 vs. 0.523). The Query-Radar attention mask can prevent attentions for certain pairs of queries and radar features based on their distances. Without it, each query has to interact with all the radar features (1500 in our work) within the scene. This is challenging for the network to fuse useful radar features with the query, resulting in poorer performance. 

\textbf{The iterative refinement:} There are three transformer cross-attention decoders that work iteratively in TransCAR. We study the effectiveness of the iterative design in TransCAR fusion and present the results in Table \ref{nuscenes_fusion_layer_table}.  The quantitative results in Table \ref{nuscenes_fusion_layer_table} suggests that the iterative refinement in TransCAR fusion can improve the detection performance and is beneficial to fully leverage our proposed fusion architecture.

\definecolor{LightCyan}{RGB}{153,204,255}
\begin{table*}
\setlength{\tabcolsep}{1.5pt}
\centering
%\resizebox{0.8\textwidth}{!}
\begin{center}
\begin{tabular}{|c|c|c|c|c|c|c|c|c|c|c|c|c|}
\hline
\hline
\multirow{2}{*}{\footnotesize{Method}} & \multicolumn{3}{c|}{\makecell{\footnotesize{All $\leq$20m}}} & \multicolumn{3}{c|}{\makecell{\footnotesize{All in 20 - 30m}}} & \multicolumn{3}{c|}{\makecell{\footnotesize{All in 30 - 40m}}} & \multicolumn{3}{c|}{\makecell{\footnotesize{All in 40 - 50m}}} \\\cline{2-13} & \footnotesize{mAP$\uparrow$} & \footnotesize{NDS$\uparrow$} & \footnotesize{mAVE$\downarrow$} & \footnotesize{mAP$\uparrow$} & \footnotesize{NDS$\uparrow$} & \footnotesize{mAVE$\downarrow$} & \footnotesize{mAP$\uparrow$} & \footnotesize{NDS$\uparrow$} & \footnotesize{mAVE$\downarrow$} & \footnotesize{mAP$\uparrow$} & \footnotesize{NDS$\uparrow$} & \footnotesize{mAVE$\downarrow$} \\
\hline
\footnotesize{DETR3D \cite{wang2022detr3d}} & \footnotesize{{\space\space47.9\space\space}} & \footnotesize{48.9} & \footnotesize{0.996} & \footnotesize{{\space\space25.8\space\space}} & \footnotesize{38.4} & \footnotesize{0.722} & \footnotesize{{\space\space11.0\space\space}} & \footnotesize{23.6} & \footnotesize{0.997} & \footnotesize{{\space\space0.9\space\space}} & \footnotesize{10.0} & \footnotesize{1.098}\\
\hline
\footnotesize{TransCAR (Ours)} & \footnotesize{48.6} & \footnotesize{54.0} & \footnotesize{0.537} & \footnotesize{26.6} & \footnotesize{41.9} & \footnotesize{0.450} & \footnotesize{11.9} & \footnotesize{27.4} & \footnotesize{0.705} & \footnotesize{1.0} & \footnotesize{10.8} & \footnotesize{0.938}\\
\hline
%\rowcolor{LightCyan}
\footnotesize{Improvement} & \footnotesize{+0.7} & \footnotesize{+5.1} & \footnotesize{-0.459} & \footnotesize{+0.8} & \footnotesize{+3.5} & \footnotesize{-0.272} & \footnotesize{+0.9} & \footnotesize{+3.8} & \footnotesize{-0.292} & \footnotesize{+0.1} & \footnotesize{+0.8} & \footnotesize{-0.160}  \\
\hline
\hline
\end{tabular}
\end{center}
\caption{Mean Average Precision (mAP, $\%$), nuScenes Detection Score (NDS) and mean Average Velocity Error (AVE, $m/s$) for all classes of different distance ranges on nuScenes validation set. Our TransCAR outperforms the baseline (DETR3D) in all distance ranges.}
% \vspace{-0.25cm}
\label{TransCAR_distance_all}
\end{table*}

\textbf{Performance in different distance ranges:} Table \ref{TransCAR_distance_all} and Table \ref{TransCAR_distance_car} show the detection performance on nuScenes dataset in different distance ranges, Table \ref{TransCAR_distance_all} shows the average results for all the 10 classes, and Table \ref{TransCAR_distance_car} is for Car class only. The results from these two Tables suggest that the vision-only baseline method (DETR3D) and our TransCAR perform better in shorter distances. The improvements of TransCAR are more significant in the range of $20 - 40$ meters. This is mainly because for objects within 20 meters, the position errors are smaller, there are limited space for leveraging radar for improvement. And for objects beyond 40 meters, the baseline performs poorly, therefore TransCAR can only provided limited improvement. Note that the mean average precision (mAP) and corresponding improvements for all 10 classes in Table \ref{TransCAR_distance_all} are smaller than the ones for Car class in Table \ref{TransCAR_distance_car}. There are mainly two reasons for this. First, mAP is the mean of APs of all classes, in nuScenes dataset, radar sensor has a higher miss rate for small-sized object classes (ped, cyclist, traffic cone, etc.). For example, $78.16\%$ of pedestrians and $63.74\%$ of cyclists do not have radar returns. Therefore, the performances for these classes are worse compared to large-sized objects (car, bus, etc.).  Therefore, the improvements brought by TransCAR for these classes are limited, for those classes of objects that have radar returns, TransCAR can leverage the radar signal to improve the detection performance, for the ones that do not have radar returns, TransCAR can only preserve the baseline performance. Second, there is a significant class imbalance in the nuScenes dataset, class Car accounts for $43.74\%$ of the training instances, while for some other classes, such as class Cyclist and class Motorcycle only accounts for $1.00\%$ and $1.07\%$ respectively. The training examples for these rare classes are not sufficient. As for the major class Car, which is also the most common objects in the driving scenarios, TransCAR can improve the detection performance and velocity estimation by a large margin (Table \ref{TransCAR_distance_car}).

\definecolor{LightCyan}{RGB}{153,204,255}
\begin{table*}
\setlength{\tabcolsep}{1.5pt}
\centering
%\resizebox{0.65\textwidth}{!}
{
%\begin{center}
\begin{tabular}{|c|c|c|c|c|c|c|c|c|}
\hline
\hline
\multirow{2}{*}{\footnotesize{Method}} & \multicolumn{2}{c|}{\makecell{\footnotesize{Cars $\leq$20m} \\ \makecell{\footnotesize{Radar Miss 31.53\%}}}} & \multicolumn{2}{c|}{\makecell{\footnotesize{Cars in 20 - 30m} \\ \footnotesize{Radar Miss 73.29\%}}} & \multicolumn{2}{c|}{\makecell{\footnotesize{Cars in 30 - 40m}\\\footnotesize{Radar Miss 48.16\%}}} & \multicolumn{2}{c|}{\makecell{\footnotesize{Cars in 40 - 50m} \\\footnotesize{Radar Miss 50.45\%}}} \\\cline{2-9} & \footnotesize{AP$\uparrow$} & \footnotesize{AVE$\downarrow$} & \footnotesize{AP$\uparrow$} & \footnotesize{AVE$\downarrow$} & \footnotesize{AP$\uparrow$} & \footnotesize{AVE$\downarrow$} & \footnotesize{AP$\uparrow$} & \footnotesize{AVE$\downarrow$} \\
\hline
\footnotesize{DETR3D \cite{wang2022detr3d}} & \footnotesize{\space\space76.4\space\space} & \footnotesize{0.917} & \footnotesize{\space\space48.7\space\space} & \footnotesize{0.810} & \footnotesize{\space\space28.8\space\space} & \footnotesize{0.934} & \footnotesize{\space\space5.0\space\space} & \footnotesize{1.015} \\
\hline
\footnotesize{TransCAR (Ours)} &\footnotesize{79.2} & \footnotesize{0.487} & \footnotesize{53.8} & \footnotesize{0.398} & \footnotesize{33.9} & \footnotesize{0.588} & \footnotesize{6.0} & \footnotesize{0.698} \\
\hline
%\rowcolor{LightCyan}
\footnotesize{Improvement} & \footnotesize{+2.8} & \footnotesize{-0.430} & \footnotesize{+5.1} & \footnotesize{-0.412} & \footnotesize{+5.1} & \footnotesize{-0.346} & \footnotesize{+1.0} & \footnotesize{-0.317}  \\
\hline
\hline
\end{tabular}
%\end{center}
}
\caption{Average Precision (AP, $\%$) and Average Velocity Error (AVE, $m/s$) for Car class of different distance ranges on nuScenes validation set. We also present the miss rate of radar sensor for different distance ranges. A car missed by radar is defined as a car that does not have radar return. Our TransCAR improves the AP and reduces the velocity estimation error by a large margin in all distance ranges.}
% \vspace{-0.5cm}
\label{TransCAR_distance_car}
\end{table*}

\begin{comment}
\begin{figure*}
\centering
   \includegraphics[width=0.95\textwidth]{rain_study/rain_and_no_rain_0818.jpg}

   \caption{Comparison of Average Precision (AP) and Average Velocity Error (AVE, m/sm/s) for class Car under the rain and no-rain scenes on nuScenes validation set. There are 1088 frames (27 scenes) among 6019 frames (150 scenes) in nuScenes validation set are annotated as rain. TransCAR can significant improve the detection performance and reduce the velocity estimation error under rainy conditions.}
\label{rain_figure}
\end{figure*}
\end{comment}
\vspace{-0.1cm}
\textbf{Different weather and lighting conditions:} Radar is more robust under different weather and light conditions compared to cameras. We evaluate the detection performance under rainy conditions and during the night, the results are shown in Table \ref{rain_no_rain} and Table \ref{night_and_day}. Note that nuScenes does not provide the weather label for each annotated frame, the weather information is provided in the scene description section (a scene is a 20-second data segment \cite{caesar2020nuscenes}). After manual check of some of the annotated frames, we found that not all the frames under `rain' were captured when the rain was falling, some of them were collected right before or after the rain. However, these images are of lower quality compared to the ones collected during sunny weather. Therefore, they are suitable for our evaluation experiments. 

Table \ref{rain_no_rain} shows the AP and AVE for class Car under rain and no-rain scenes. TransCAR has a higher AP improvement (+5.1\% vs. +3.6\%) for the rain scenes compared to no-rain scenes. The AVE for rain scenes is smaller than in the no-rain scenes; this is because there are biases in the rain frames, and the cars within these rain scenes are closer to the ego vehicle, which makes them easier to be detected.

Table \ref{night_and_day} shows the comparison of the detection performance of night and daytime scenes. Poor lighting conditions at night make the baseline method perform worse than daytime (52.2\% vs. 54.8\% in AP, 1.691m/s vs. 0.866m/s in AVE), TransCAR can leverage the radar data and boost the AP by 6.9\% and reduce AVE by 1.012m/s. Although there are limited night scenes (15 scenes, 602 frames), this result can still demonstrate the effectiveness of TransCAR in night scenarios.

\begin{comment}
\begin{figure*}
\centering
   \includegraphics[width=0.95\textwidth]{night_and_day/night_and_day_0818.jpg}

   \caption{Comparison of Average Precision (AP) and Average Velocity Error (AVE, m/sm/s) for class Car during the night and daytime scenes on nuScenes validation set. There are 602 frames among 6019 frames in nuScenes validation set are collected during the night. TransCAR can leverage the radar data to significantly improve the performance and reduce the velocity estimation error during the night when the camera is affected.}
\label{night_figure}
\end{figure*}
\end{comment}

\definecolor{LightCyan}{RGB}{153,204,255}
\begin{table}
\setlength{\tabcolsep}{1.5pt}
\centering
%\resizebox{0.65\textwidth}{!}
{
%\begin{center}
\begin{tabular}{|c|c|c|c|c|}
\hline
\hline
\multirow{2}{*}{\footnotesize{Method}} & \multicolumn{2}{c|}{\makecell{\footnotesize{Rain}}} & \multicolumn{2}{c|}{\makecell{\footnotesize{No Rain}}} \\\cline{2-5} & \footnotesize{AP$\uparrow$} & \footnotesize{AVE$\downarrow$} & \footnotesize{AP$\uparrow$} & \footnotesize{AVE$\downarrow$} \\
\hline
\footnotesize{DETR3D \cite{wang2022detr3d}} & \footnotesize{\space\space54.2\space\space} & \footnotesize{0.675} & \footnotesize{\space\space54.7\space\space} & \footnotesize{0.965}\\
\hline
\footnotesize{TransCAR (Ours)} &\footnotesize{59.3} & \footnotesize{0.409} & \footnotesize{58.3} & \footnotesize{0.534}\\
\hline
%\rowcolor{LightCyan}
\footnotesize{Improvement} & \footnotesize{+5.1} & \footnotesize{-0.266} & \footnotesize{+3.6} & \footnotesize{-0.431}\\
\hline
\hline
\end{tabular}
%\end{center}
}
\caption{Comparison of Average Precision (AP) and Average Velocity Error (AVE, $m/s$) for class Car under the rain and no-rain scenes on nuScenes validation set. There are 1088 frames (27 scenes) among 6019 frames (150 scenes) in nuScenes validation set are annotated as rain. TransCAR can significant improve the detection performance and reduce the velocity estimation error under rainy conditions.}
% \vspace{-0.25cm}
\label{rain_no_rain}
\end{table}

\definecolor{LightCyan}{RGB}{153,204,255}
\begin{table}
\setlength{\tabcolsep}{1.5pt}
\centering
%\resizebox{0.65\textwidth}{!}
{
%\begin{center}
\begin{tabular}{|c|c|c|c|c|}
\hline
\hline
\multirow{2}{*}{\footnotesize{Method}} & \multicolumn{2}{c|}{\makecell{\footnotesize{Night}}} & \multicolumn{2}{c|}{\makecell{\footnotesize{Daytime}}} \\\cline{2-5} & \footnotesize{AP$\uparrow$} & \footnotesize{AVE$\downarrow$} & \footnotesize{AP$\uparrow$} & \footnotesize{AVE$\downarrow$} \\
\hline
\footnotesize{DETR3D \cite{wang2022detr3d}} & \footnotesize{\space\space52.2\space\space} & \footnotesize{1.691} & \footnotesize{\space\space54.8\space\space} & \footnotesize{0.866}\\
\hline
\footnotesize{TransCAR (Ours)} &\footnotesize{59.1} & \footnotesize{0.679} & \footnotesize{58.4} & \footnotesize{0.500}\\
\hline
%\rowcolor{LightCyan}
\footnotesize{Improvement} & \footnotesize{+6.9} & \footnotesize{-1.012} & \footnotesize{+3.6} & \footnotesize{-0.366}\\
\hline
\hline
\end{tabular}
%\end{center}
}
\caption{Comparison of Average Precision (AP) and Average Velocity Error (AVE, $m/s$) for class Car during the night and daytime scenes on nuScenes validation set. There are 602 frames among 6019 frames in nuScenes validation set are collected during the night. TransCAR can leverage the radar data to significantly improve the performance and reduce the velocity estimation error during the night when the camera is affected.}
\vspace{-0.5cm}
\label{night_and_day}
\end{table}

\section{Supplementary Materials}

\subsection{Comparison of Radar and LiDAR Miss Rate}
Compared to LiDAR, Radar has much higher miss rate. There are mainly two reasons: First, Automotive radar has a very limited field of view compared to LiDAR. And radar is usually installed at a lower position. Therefore, any object that is located out of the radar's small vertical field of view will be missed. Second, radar beams are wider and the azimuth resolution is limited, making it difficult to detect small objects. The statistics of LiDAR and radar miss rate for nuScenes training set are shown in Table~\ref{nuscenes_miss_stat}. We counted the number of objects in different classes, the number of objects missed by radar or LiDAR and the corresponding miss rate. Radar has a high miss rate, especially for small objects. For the two most common classes on the road, car and pedestrian, radar misses 36.05\% of cars and 78.16\% of pedestrians. 

\begin{table*}[t]
\centering
\resizebox{0.8\textwidth}{!}
{
\begin{tabular}{|c|c|c|c|c|c|}
\hline
\hline
\footnotesize{Class} & \footnotesize{\#Total} & \footnotesize{\#Radar Misses} & \footnotesize{Radar Miss Rate} & \footnotesize{\#LiDAR Misses} & \footnotesize{LiDAR Miss Rate} \\
\hline
\footnotesize{Car} &  \footnotesize{318157} & \footnotesize{114697} & \footnotesize{36.05\%} & \footnotesize{14196} & \footnotesize{4.46\%} \\
\hline
\footnotesize{Truck} & \footnotesize{47687} & \footnotesize{12779} & \footnotesize{26.80\%} & \footnotesize{1134} & \footnotesize{2.38\%}  \\
\hline
\footnotesize{Bus} & \footnotesize{7451} & \footnotesize{1521} & \footnotesize{20.41\%} & \footnotesize{42} & \footnotesize{0.56\%}  \\
\hline
\footnotesize{Trailer} & \footnotesize{12604} & \footnotesize{2413} & \footnotesize{19.14\%} & \footnotesize{413} & \footnotesize{3.28\%}  \\
\hline
\footnotesize{C.V.*} & \footnotesize{8655} & \footnotesize{2611} & \footnotesize{30.17\%} & \footnotesize{166} & \footnotesize{1.92\%}  \\
\hline
\footnotesize{Ped.*} & \footnotesize{139493} & \footnotesize{109026} & \footnotesize{78.16\%} & \footnotesize{473} & \footnotesize{0.34\%}  \\
\hline
\footnotesize{Motor.*} & \footnotesize{9209} & \footnotesize{5197} & \footnotesize{56.43\%} & \footnotesize{196} & \footnotesize{2.13\%}  \\
\hline
\footnotesize{Bike} & \footnotesize{8171} & \footnotesize{5208} & \footnotesize{63.74\%} & \footnotesize{111} & \footnotesize{1.36\%}  \\
\hline
\footnotesize{T.C.*} & \footnotesize{78532} & \footnotesize{54622} & \footnotesize{69.55\%} & \footnotesize{1140} & \footnotesize{1.45\%}  \\
\hline
\footnotesize{Barrier} & \footnotesize{115105} & \footnotesize{81464} & \footnotesize{70.77\%} & \footnotesize{1828} & \footnotesize{1.59\%}  \\
\hline
\hline
\end{tabular}
}
\caption{Statistics of objects in different classes in nuScenes training set within 50 meters of the ego vehicle. * `C.V.', `Ped', `Motor' and `T.C' represent construction vehicle, pedestrian, motorcycle and traffic cone, respectively. An object that is missed by radar or LiDAR is defined as having no hit/return from that object. Radar misses more objects. For the two most common classes in autonomous driving applications, car and pedestrian, radar misses 36.05\% cars and 78.16\% pedestrians. Although nuScenes does not provide detailed visibilities of objects in the image, we believe that it is much higher than radar. Therefore, we use camera instead of radar to generate 3D object queries for fusion. }
\label{nuscenes_miss_stat}
\end{table*}

Note that we are not criticising radar. Table~\ref{nuscenes_miss_stat} shows the challenges of using radar for object detection tasks. Understanding the properties of radar measurements help us to design a reasonable fusion system. As discussed in the main paper, based on these statistics, we conclude that radar, as configured on the nuScenes vehicle, is not suitable to be used to generate the 3D queries. 

We note that despite these physical limitations on the radar sensors, our results show that fusion with radar can significantly improve image-only detection.  This opens the possibility of configuring radar differently, such as on the vehicle roof, to reduce the miss rate and potentially improve fusion performance further.

\begin{table*}[t]
%\resizebox{\textwidth}{!}{
\setlength{\tabcolsep}{1.0pt}
\renewcommand{\arraystretch}{1.1} 
\begin{center}

\begin{tabular}{|c|c|c|c|c|c|c|c|c|}
\hline
\hline
{Method} & {Sensor*} & {mAP$\uparrow$} & {NDS$\uparrow$} & {mATE$\downarrow$} & {mAVE$\downarrow$} & {mASE$\downarrow$} & {mAOE$\downarrow$} & {mAAE$\downarrow$} \\ 
\hline
{MonoDIS\cite{simonelli2019disentangling}}& {C} & {30.4} & {38.4} & {0.738} & {1.553} & {0.263} & {0.546} & {0.134} \\

{CenterNet\cite{zhou2019objects}}& {C} & {33.8} & {40.0} & {0.658} & {1.629} & {0.255} & {0.629} & {0.142} \\
%\hline
{FCOS3D\cite{wang2021fcos3d}} & {C} & {35.8} & {42.8} & {0.690} & {1.434} & {0.249} & {0.452} & {0.124} \\
{PGD\cite{wang2022probabilistic}} & {C} & {38.6} & {44.8} & \textbf{{0.626}} & {1.509} & \textbf{0.245} & {0.451} & {0.127} \\
{DETR3D\cite{wang2022detr3d}(baseline)} & {C} & {41.2} & {47.9} & {0.641} & {0.845} & {0.255} & {0.394} & {0.133} \\
\specialrule{1pt}{0pt}{0pt}
{PointPillar \cite{lang2019pointpillars}}& {L} & {30.5} & {45.3} & {0.517} & {0.316} & {0.290} & {0.500} & {0.368} \\
%\hline
{infoFocus\cite{wang2020infofocus}}& {L} & {39.5} & {39.5} & {0.363} & {1.000} & {0.265} & {1.132} & {0.395} \\
%\hline
%{  PointPainting \cite{Vora_2020_CVPR}}& {CL} & {  40.1} & { 55.0} & { 0.392} & { 0.270} & { 0.269} & { 0.476} & { 0.270} \\
\specialrule{1pt}{0pt}{0pt}
{CenterFusion\cite{nabati2021centerfusion}} & {CR} & {32.6} & {44.9} & {0.631} & {0.614} & {0.261} & {0.516} & \textbf{0.115} \\
%\hline
{TransCAR(Ours)} & {CR} & {\textbf{42.2}} & \textbf{52.2} & {0.630} & \textbf{0.495} & {0.260} & \textbf{0.384} & {0.121} \\

\hline
\hline
\end{tabular}
\begin{tablenotes}
            \small
            \item *In Sensor column, C represents Camera-only. L stands for LiDAR-only, CR shows Camera and Radar fusion-based approaches.
        \end{tablenotes}
        %\vspace{-0.5cm}
\end{center}
%\vspace{-0.5cm}
\caption{Comparison with other SOTA methods in other secondary evaluation metrics defined by nuScenes dataset on nuScenes test set.  The best performers are
highlighted in bold exclude LiDAR-only solutions} 
\label{muscenes_second_test_table}
\end{table*}

\subsection{More Results on Secondary Evaluation Metrics}
For completeness, we present Table \ref{muscenes_second_test_table} to show the comparison with other SOTA methods in other secondary evaluation metrics on the nuScenes test set. After fusing with radar, compared with other methods, our TransCAR has either better or same level of performance in all of the secondary evaluation metrics. In particular, with the fusion of radar signals, TransCAR outperform other methods in velocity estimation by a large margin. Compared to baseline (DETR3D), TransCAR improves the performance in all evaluation metrics.

\subsection{More Ablation Studies}
\textbf{Number of radar frames for fusion:} We accumulate multiple radar frames with ego vehicle motion compensation for fusion as radar point clouds are sparse. We evaluate the impact of accumulating different number of radar frames, and the results are presented in Table~\ref{radar_frames_all} and Table~\ref{radar_frames_car}. Table~\ref{radar_frames_all} shows the evaluation results for all 10 classes in the nuScenes validation set, and Table~\ref{radar_frames_car} is the result for car class only. Accumulating 5 radar frames provides the overall best results, while the velocity error mAVE is slightly lower for accumulating 10 frames. Accumulating more radar frames can give us a denser radar point cloud for fusion, but at the same time, more noise points will be included. Also, for fast moving objects, accumulating more radar frames will generate a longer ``trail'' as the object motion cannot be compensated at this stage. This could potentially harm the bounding box location estimation.

\begin{table}
%\resizebox{\textwidth}{!}{
\setlength{\tabcolsep}{1.0pt}
\renewcommand{\arraystretch}{1.1} 
\begin{center}

\begin{tabular}{|c|c|c|c|c|}

\hline
\small\makecell{Number of accumulated \\ radar frames}  & {\small mAP$\uparrow$} & {\small NDS$\uparrow$} & {\small mATE$\downarrow$} & {\small mAVE$\downarrow$}\\ 
\hline
\makecell{\small 0 (Baseline\cite{wang2022detr3d})} & {\small34.6} & {\small42.2} & {\small0.823} & {\small 0.876}\\
\hline
\makecell{\small 1} & {\small 35.3} & {\small 44.3} & {\small 0.762} & {\small 0.703}\\
\hline
\makecell{\small{3}}  & \small{35.4} & \small{44.9} & \small{\textbf{0.755}} & \small{0.657} \\%& {55.9} & {29.4} & {34.8} & \textbf{\color{gg}16.7} & {8.6} & {42.6} & {34.1} & {28.3} & {52.8} & \textbf{\color{gg}47.0} \\
\hline
{\small{5}}  & \small{\textbf{35.5}} & \small{\textbf{46.4}} & \small{0.759} & \small{0.523} \\
\hline
{\small{10}}  & \small{35.4} & \small{46.3} & \small{0.764} & \small{\textbf{0.521}} \\

\hline
\hline
\end{tabular}
\end{center}
\caption{Evaluation on \textit{all classes} detection results from accumulating different number of radar frames for fusion.}
\label{radar_frames_all}
\end{table}

\begin{table}
%\resizebox{\textwidth}{!}{
\setlength{\tabcolsep}{1.0pt}
\renewcommand{\arraystretch}{1.1} 
\begin{center}

\begin{tabular}{|c|c|c|c|}

\hline
\small\makecell{Number of accumulated \\ radar frames}  & {\small mAP$\uparrow$} & {\small mATE$\downarrow$} & {\small mAVE$\downarrow$}\\ 
\hline
\makecell{\small 0 (Baseline\cite{wang2022detr3d})} & {\small54.63} & {\small0.544} & {\small 0.911}\\
\hline
\makecell{\small 1} & {\small 57.6} & {\small 0.508} & {\small 0.729}\\
\hline
\makecell{\small{3}}  & \small{58.26} & \small{0.494} & \small{0.641} \\%& {55.9} & {29.4} & {34.8} & \textbf{\color{gg}16.7} & {8.6} & {42.6} & {34.1} & {28.3} & {52.8} & \textbf{\color{gg}47.0} \\
\hline
{\small{5}}  & \small{\textbf{58.52}} & \small{\textbf{0.493}} & \small{0.511} \\
\hline
{\small{10}}  & \small{58.37} & \small{0.496} & \small{\textbf{0.501}} \\

\hline
\hline
\end{tabular}
\end{center}
\caption{Evaluation on \textit{Car} detection results from accumulating different number of radar frames for fusion.}
\label{radar_frames_car}
\end{table}

\textbf{Attention mask radii:} In our TransCAR fusion system, we apply circle attention masks for transformer decoders. The radii of the circle attention masks can be different for each transformer decoder. We test different circle attention mask radius combinations on the nuScenes validation set, and the results are shown in Table~\ref{mask_radius_study}. As shown in the Table~\ref{mask_radius_study}, all radius configurations can improve the 3D detection performance significantly. There is a trade-off between large and small attention masks. Large attention mask can increase the probability of incorporating the right radar features, especially for large objects, but the chances of including noise and radar features from other nearby objects are also higher. As for small attention mask, we are more certain the included radar features are good ones, but there is a higher chance that we may miss the right radar features. According to Table \ref{mask_radius_study}, $(2m,2m,1m)$ has the best overall performance. Larger radii at the beginning provides a higher chance for detections to capture right radar features. A smaller radius at the end is better for the final estimation refinement.

\begin{table}
%\resizebox{\textwidth}{!}{
\setlength{\tabcolsep}{1.0pt}
\renewcommand{\arraystretch}{1.1} 
\begin{center}
\begin{tabular}{|c|c|c|c|c|}

\hline
\small\makecell{Attention mask radii for\\ 3 transformer decoders}  & {\small mAP$\uparrow$} & {\small NDS$\uparrow$} & {\small mATE$\downarrow$} & {\small mAVE$\downarrow$}\\ 
\hline
\makecell{\small Baseline\cite{wang2022detr3d}} & {\small34.6} & {\small42.2} & {\small0.823} & {\small 0.876}\\
\hline
\makecell{\small ($2m,2m,2m$)}  & \small{35.2} & \small{\textbf{46.4}} & \small{{0.761}} & \small{\textbf{0.512}} \\%& {55.9} & {29.4} & {34.8} & \textbf{\color{gg}16.7} & {8.6} & {42.6} & {34.1} & {28.3} & {52.8} & \textbf{\color{gg}47.0} \\
\hline
{\small{($2m,2m,1m$)}}  & \small{\textbf{35.5}} & \small{\textbf{46.4}} & \small{0.759} & \small{0.523} \\
\hline
\makecell{\small ($3m,2m,1m$)} & {\small 35.3} & {\small 45.1} & {\small 0.759} & {\small 0.630}\\
\hline
{\small{($2m,1m,1m$)}}  & \small{35.4} & \small{46.2} & \small{\textbf{0.757}} & \small{{0.541}} \\
\hline
{\small{($1m,1m,1m$)}}  & \small{35.4} & \small{45.1} & \small{0.758} & \small{{0.643}} \\

\hline
\hline
\end{tabular}
\end{center}
\caption{Comparison of different attention mask radii for each transformer decoder in our TransCAR.}
\label{mask_radius_study}
\end{table}

% \vspace{-0.2cm}
\section{Conclusion}
In this paper, we proposed TransCAR, an effective and robust Transformer-based Camera-And-Radar 3D detection framework that can learn a soft-association between radar features and vision queries instead of a hard-association based on sensor calibration. The associated radar features improve range and velocity estimation. Our TransCAR sets the new state-of-the-art camera-radar detection performance on the challenging nuScenes detection benchmark. We hope that our work can inspire further research on radar-camera fusion and motivate using transformers for sensor fusion.

%%%%%%%%% REFERENCES
\newpage
{\small
\bibliographystyle{ieee_fullname}
\bibliography{egbib}

\begin{thebibliography}{10}\itemsep=-1pt

\bibitem{bai2022transfusion}
Xuyang Bai, Zeyu Hu, Xinge Zhu, Qingqiu Huang, Yilun Chen, Hongbo Fu, and
  Chiew-Lan Tai.
\newblock Transfusion: Robust lidar-camera fusion for 3d object detection with
  transformers.
\newblock In {\em Proceedings of the IEEE/CVF Conference on Computer Vision and
  Pattern Recognition}, pages 1090--1099, 2022.

\bibitem{brazil2019m3d}
Garrick Brazil and Xiaoming Liu.
\newblock M3d-rpn: Monocular 3d region proposal network for object detection.
\newblock In {\em Proceedings of the IEEE International Conference on Computer
  Vision}, pages 9287--9296, 2019.

\bibitem{caesar2020nuscenes}
Holger Caesar, Varun Bankiti, Alex~H Lang, Sourabh Vora, Venice~Erin Liong,
  Qiang Xu, Anush Krishnan, Yu Pan, Giancarlo Baldan, and Oscar Beijbom.
\newblock nuscenes: A multimodal dataset for autonomous driving.
\newblock In {\em Proceedings of the IEEE/CVF conference on computer vision and
  pattern recognition}, pages 11621--11631, 2020.

\bibitem{carion2020end}
Nicolas Carion, Francisco Massa, Gabriel Synnaeve, Nicolas Usunier, Alexander
  Kirillov, and Sergey Zagoruyko.
\newblock End-to-end object detection with transformers.
\newblock In {\em European conference on computer vision}, pages 213--229.
  Springer, 2020.

\bibitem{chang2019argoverse}
Ming-Fang Chang, John Lambert, Patsorn Sangkloy, Jagjeet Singh, Slawomir Bak,
  Andrew Hartnett, De Wang, Peter Carr, Simon Lucey, Deva Ramanan, et~al.
\newblock Argoverse: 3d tracking and forecasting with rich maps.
\newblock In {\em Proceedings of the IEEE Conference on Computer Vision and
  Pattern Recognition}, pages 8748--8757, 2019.

\bibitem{chen2017multi}
Xiaozhi Chen, Huimin Ma, Ji Wan, Bo Li, and Tian Xia.
\newblock Multi-view 3d object detection network for autonomous driving.
\newblock In {\em Proceedings of the IEEE Conference on Computer Vision and
  Pattern Recognition}, pages 1907--1915, 2017.

\bibitem{chen2020monopair}
Yongjian Chen, Lei Tai, Kai Sun, and Mingyang Li.
\newblock Monopair: Monocular 3d object detection using pairwise spatial
  relationships.
\newblock In {\em Proceedings of the IEEE/CVF Conference on Computer Vision and
  Pattern Recognition}, pages 12093--12102, 2020.

\bibitem{dosovitskiy2020image}
Alexey Dosovitskiy, Lucas Beyer, Alexander Kolesnikov, Dirk Weissenborn,
  Xiaohua Zhai, Thomas Unterthiner, Mostafa Dehghani, Matthias Minderer, Georg
  Heigold, Sylvain Gelly, et~al.
\newblock An image is worth 16x16 words: Transformers for image recognition at
  scale.
\newblock {\em arXiv preprint arXiv:2010.11929}, 2020.

\bibitem{geiger2012we}
Andreas Geiger, Philip Lenz, and Raquel Urtasun.
\newblock Are we ready for autonomous driving? the kitti vision benchmark
  suite.
\newblock In {\em 2012 IEEE Conference on Computer Vision and Pattern
  Recognition}, pages 3354--3361. IEEE, 2012.

\bibitem{he2016deep}
Kaiming He, Xiangyu Zhang, Shaoqing Ren, and Jian Sun.
\newblock Deep residual learning for image recognition.
\newblock In {\em Proceedings of the IEEE conference on computer vision and
  pattern recognition}, pages 770--778, 2016.

\bibitem{huang2020epnet}
Tengteng Huang, Zhe Liu, Xiwu Chen, and Xiang Bai.
\newblock Epnet: Enhancing point features with image semantics for 3d object
  detection.
\newblock In {\em European Conference on Computer Vision}, pages 35--52.
  Springer, 2020.

\bibitem{ku2018joint}
Jason Ku, Melissa Mozifian, Jungwook Lee, Ali Harakeh, and Steven~L Waslander.
\newblock Joint 3d proposal generation and object detection from view
  aggregation.
\newblock In {\em 2018 IEEE/RSJ International Conference on Intelligent Robots
  and Systems (IROS)}, pages 1--8. IEEE, 2018.

\bibitem{kuhn1955hungarian}
Harold~W Kuhn.
\newblock The hungarian method for the assignment problem.
\newblock {\em Naval research logistics quarterly}, 2(1-2):83--97, 1955.

\bibitem{lang2019pointpillars}
Alex~H Lang, Sourabh Vora, Holger Caesar, Lubing Zhou, Jiong Yang, and Oscar
  Beijbom.
\newblock Pointpillars: Fast encoders for object detection from point clouds.
\newblock {\em Proceedings of the IEEE Conference on Computer Vision and
  Pattern Recognition}, 2019.

\bibitem{liang2019multi}
Ming Liang, Bin Yang, Yun Chen, Rui Hu, and Raquel Urtasun.
\newblock Multi-task multi-sensor fusion for 3d object detection.
\newblock In {\em Proceedings of the IEEE Conference on Computer Vision and
  Pattern Recognition}, pages 7345--7353, 2019.

\bibitem{liang2018deep}
Ming Liang, Bin Yang, Shenlong Wang, and Raquel Urtasun.
\newblock Deep continuous fusion for multi-sensor 3d object detection.
\newblock In {\em Proceedings of the European Conference on Computer Vision
  (ECCV)}, pages 641--656, 2018.

\bibitem{lin2017feature}
Tsung-Yi Lin, Piotr Doll{\'a}r, Ross Girshick, Kaiming He, Bharath Hariharan,
  and Serge Belongie.
\newblock Feature pyramid networks for object detection.
\newblock In {\em Proceedings of the IEEE Conference on Computer Vision and
  Pattern Recognition}, pages 2117--2125, 2017.

\bibitem{lin2017focal}
Tsung-Yi Lin, Priya Goyal, Ross Girshick, Kaiming He, and Piotr Doll{\'a}r.
\newblock Focal loss for dense object detection.
\newblock In {\em Proceedings of the IEEE international conference on computer
  vision}, pages 2980--2988, 2017.

\bibitem{mousavian20173d}
Arsalan Mousavian, Dragomir Anguelov, John Flynn, and Jana Kosecka.
\newblock 3d bounding box estimation using deep learning and geometry.
\newblock In {\em Proceedings of the IEEE Conference on Computer Vision and
  Pattern Recognition}, pages 7074--7082, 2017.

\bibitem{nabati2021centerfusion}
Ramin Nabati and Hairong Qi.
\newblock Centerfusion: Center-based radar and camera fusion for 3d object
  detection.
\newblock In {\em Proceedings of the IEEE/CVF Winter Conference on Applications
  of Computer Vision}, pages 1527--1536, 2021.

\bibitem{pang2020clocs}
Su Pang, Daniel Morris, and Hayder Radha.
\newblock Clocs: Camera-lidar object candidates fusion for 3d object detection.
\newblock {\em 2020 IEEE/RSJ International Conference on Intelligent Robots and
  Systems (IROS)}, 2020.

\bibitem{pang2022fast}
Su Pang, Daniel Morris, and Hayder Radha.
\newblock Fast-clocs: Fast camera-lidar object candidates fusion for 3d object
  detection.
\newblock In {\em Proceedings of the IEEE/CVF Winter Conference on Applications
  of Computer Vision}, pages 187--196, 2022.

\bibitem{qi2018frustum}
Charles~R Qi, Wei Liu, Chenxia Wu, Hao Su, and Leonidas~J Guibas.
\newblock Frustum pointnets for 3d object detection from rgb-d data.
\newblock In {\em Proceedings of the IEEE Conference on Computer Vision and
  Pattern Recognition}, pages 918--927, 2018.

\bibitem{CaDDN}
Cody Reading, Ali Harakeh, Julia Chae, and Steven~L. Waslander.
\newblock Categorical depth distributionnetwork for monocular 3d object
  detection.
\newblock {\em CVPR}, 2021.

\bibitem{shi2020pv}
Shaoshuai Shi, Chaoxu Guo, Li Jiang, Zhe Wang, Jianping Shi, Xiaogang Wang, and
  Hongsheng Li.
\newblock Pv-rcnn: Point-voxel feature set abstraction for 3d object detection.
\newblock In {\em CVPR}, 2020.

\bibitem{shi2019pointrcnn}
Shaoshuai Shi, Xiaogang Wang, and Hongsheng Li.
\newblock Pointrcnn: 3d object proposal generation and detection from point
  cloud.
\newblock In {\em Proceedings of the IEEE Conference on Computer Vision and
  Pattern Recognition}, pages 770--779, 2019.

\bibitem{simonelli2019disentangling}
Andrea Simonelli, Samuel~Rota Bulo, Lorenzo Porzi, Manuel L{\'o}pez-Antequera,
  and Peter Kontschieder.
\newblock Disentangling monocular 3d object detection.
\newblock In {\em Proceedings of the IEEE/CVF International Conference on
  Computer Vision}, pages 1991--1999, 2019.

\bibitem{stewart2016end}
Russell Stewart, Mykhaylo Andriluka, and Andrew~Y Ng.
\newblock End-to-end people detection in crowded scenes.
\newblock In {\em Proceedings of the IEEE conference on computer vision and
  pattern recognition}, pages 2325--2333, 2016.

\bibitem{sun2020scalability}
Pei Sun, Henrik Kretzschmar, Xerxes Dotiwalla, Aurelien Chouard, Vijaysai
  Patnaik, Paul Tsui, James Guo, Yin Zhou, Yuning Chai, Benjamin Caine, et~al.
\newblock Scalability in perception for autonomous driving: Waymo open dataset.
\newblock In {\em Proceedings of the IEEE/CVF Conference on Computer Vision and
  Pattern Recognition}, pages 2446--2454, 2020.

\bibitem{vaswani2017attention}
Ashish Vaswani, Noam Shazeer, Niki Parmar, Jakob Uszkoreit, Llion Jones,
  Aidan~N Gomez, {\L}ukasz Kaiser, and Illia Polosukhin.
\newblock Attention is all you need.
\newblock {\em Advances in neural information processing systems}, 30, 2017.

\bibitem{Vora_2020_CVPR}
Sourabh Vora, Alex~H. Lang, Bassam Helou, and Oscar Beijbom.
\newblock Pointpainting: Sequential fusion for 3d object detection.
\newblock In {\em Proceedings of the IEEE/CVF Conference on Computer Vision and
  Pattern Recognition (CVPR)}, June 2020.

\bibitem{wang2020infofocus}
Jun Wang, Shiyi Lan, Mingfei Gao, and Larry~S Davis.
\newblock Infofocus: 3d object detection for autonomous driving with dynamic
  information modeling.
\newblock In {\em European Conference on Computer Vision}, pages 405--420.
  Springer, 2020.

\bibitem{wang2022probabilistic}
Tai Wang, ZHU Xinge, Jiangmiao Pang, and Dahua Lin.
\newblock Probabilistic and geometric depth: Detecting objects in perspective.
\newblock In {\em Conference on Robot Learning}, pages 1475--1485. PMLR, 2022.

\bibitem{wang2021fcos3d}
Tai Wang, Xinge Zhu, Jiangmiao Pang, and Dahua Lin.
\newblock Fcos3d: Fully convolutional one-stage monocular 3d object detection.
\newblock In {\em Proceedings of the IEEE/CVF International Conference on
  Computer Vision}, pages 913--922, 2021.

\bibitem{wang2022detr3d}
Yue Wang, Vitor~Campagnolo Guizilini, Tianyuan Zhang, Yilun Wang, Hang Zhao,
  and Justin Solomon.
\newblock Detr3d: 3d object detection from multi-view images via 3d-to-2d
  queries.
\newblock In {\em Conference on Robot Learning}, pages 180--191. PMLR, 2022.

\bibitem{wang2021object}
Yue Wang and Justin~M Solomon.
\newblock Object dgcnn: 3d object detection using dynamic graphs.
\newblock {\em Advances in Neural Information Processing Systems}, 34, 2021.

\bibitem{wang2019frustum}
Zhixin Wang and Kui Jia.
\newblock Frustum convnet: Sliding frustums to aggregate local point-wise
  features for amodal 3d object detection.
\newblock In {\em IROS}. IEEE, 2019.

\bibitem{xie2020pi}
Liang Xie, Chao Xiang, Zhengxu Yu, Guodong Xu, Zheng Yang, Deng Cai, and
  Xiaofei He.
\newblock Pi-rcnn: An efficient multi-sensor 3d object detector with
  point-based attentive cont-conv fusion module.
\newblock In {\em Proceedings of the AAAI Conference on Artificial
  Intelligence}, volume~34, pages 12460--12467, 2020.

\bibitem{xu2018pointfusion}
Danfei Xu, Dragomir Anguelov, and Ashesh Jain.
\newblock Pointfusion: Deep sensor fusion for 3d bounding box estimation.
\newblock In {\em Proceedings of the IEEE Conference on Computer Vision and
  Pattern Recognition}, pages 244--253, 2018.

\bibitem{yan2018second}
Yan Yan, Yuxing Mao, and Bo Li.
\newblock Second: Sparsely embedded convolutional detection.
\newblock {\em Sensors}, 18(10):3337, 2018.

\bibitem{std2019yang}
Zetong Yang, Yanan Sun, Shu Liu, Xiaoyong Shen, and Jiaya Jia.
\newblock {STD:} sparse-to-dense 3d object detector for point cloud.
\newblock {\em ICCV}, 2019.

\bibitem{yin2021center}
Tianwei Yin, Xingyi Zhou, and Philipp Kr{\"a}henb{\"u}hl.
\newblock Center-based 3d object detection and tracking.
\newblock {\em CVPR}, 2021.

\bibitem{yoo20203d}
Jin~Hyeok Yoo, Yecheol Kim, and Ji~Song Kim.
\newblock 3d-cvf: Generating joint camera and lidar features using cross-view
  spatial feature fusion for 3d object detection.

\bibitem{zhou2019objects}
Xingyi Zhou, Dequan Wang, and Philipp Kr{\"a}henb{\"u}hl.
\newblock Objects as points.
\newblock {\em arXiv preprint arXiv:1904.07850}, 2019.

\bibitem{zhou2018voxelnet}
Yin Zhou and Oncel Tuzel.
\newblock Voxelnet: End-to-end learning for point cloud based 3d object
  detection.
\newblock In {\em Proceedings of the IEEE Conference on Computer Vision and
  Pattern Recognition}, pages 4490--4499, 2018.

\bibitem{zhu2019class}
Benjin Zhu, Zhengkai Jiang, Xiangxin Zhou, Zeming Li, and Gang Yu.
\newblock Class-balanced grouping and sampling for point cloud 3d object
  detection.
\newblock {\em arXiv preprint arXiv:1908.09492}, 2019.

\bibitem{zhu2020deformable}
Xizhou Zhu, Weijie Su, Lewei Lu, Bin Li, Xiaogang Wang, and Jifeng Dai.
\newblock Deformable detr: Deformable transformers for end-to-end object
  detection.
\newblock In {\em International Conference on Learning Representations}, 2020.

\end{thebibliography}
}

\end{document}